\documentclass{article}
\usepackage[preprint]{neurips_2026}
\usepackage[utf8]{inputenc}
\usepackage[T1]{fontenc}
\usepackage{hyperref}
\hypersetup{hidelinks,hypertexnames=false}
\usepackage{url}
\usepackage{graphicx}
\usepackage{booktabs}
\usepackage{amsfonts}
\usepackage{amsmath}
\usepackage{nicefrac}
\usepackage{microtype}
\usepackage{xcolor}
\usepackage{float}
\usepackage{needspace}
\usepackage{tikz}
\usetikzlibrary{arrows.meta,positioning}
\title{Why Sub-4-Bit Vision CNNs Collapse---and Where Mixed-Precision Recovery Stops}
\author{%
  Michael Schroeder \\
  GE Appliances, a Haier company \\
  \texttt{michael.schroeder2@geappliances.com}
}

\begin{document}
\maketitle

\begin{abstract}
Compressing convolutional vision models from FP32 to 8-bit integers (INT8) is nearly lossless: on a mature EfficientNet-B0 checkpoint ($72.86\%$ ImageNet-1k top-1), full-integer INT8 LiteRT conversion shrinks model size by $70.7\%$ ($20.43\text{ MB} \rightarrow 6.00\text{ MB}$) and lowers median CPU latency by $32.2\%$ ($9.81\text{ ms} \rightarrow 6.65\text{ ms}$) with only $0.06$ percentage points accuracy loss ($72.81\%$). Pushing compression to uniform 4-bit (W4) or 3-bit (W3) weights, however, causes severe accuracy degradation to $23.03\%$ (W4) and $0.30\%$ (W3). We analyze the mechanisms of sub-4-bit degradation and establish the limits of mixed-precision recovery across all 50,000 ImageNet-1k validation images. Through a 46-arm layerwise sensitivity screen, an 82-layer dynamic-programming (DP) bit-allocation study, and a 16-arm factorial ablation crossing precision maps, per-channel MSE clipping, and BatchNorm (BN) recalibration, we trace this degradation to three factors: (1) high sensitivity in early $3\times 3$ depthwise filters, where protecting $0.34\%$ of weights ($14\text{ KB}$) at W8 restores $+42.92$ points on W4 ($65.95\%$) while FP16 adds only $+0.03$ points; (2) outlier-stretched step sizes that interact strongly with stale BN statistics ($+3.94$ to $+9.00$ points recovery); and (3) a $\sim 3.9$-point cumulative error floor at $3.56$ mean bits, where DP allocation reaches $67.94\%$ post-training and $68.98\%$ after 1-epoch QAT with BN recalibration. Crucially, CPU benchmarks expose a sharp \textbf{hardware reality gap}: whereas INT8 accelerates all 82 layers via native 8-bit SIMD instructions ($6.65\text{ ms}$), exported 4-bit (WI4, $3.59\text{ MB}$) and packed 3-bit (INT3, $2.47\text{ MB}$) models lack native sub-byte depthwise kernels on standard ARM/x86 CPUs and run $1.48\times$ ($15.43\text{ ms}$) to $21.5\times$ ($376.8\text{ ms}$) slower than FP32 due to software bit-unpacking overhead.
\end{abstract}

\section{Introduction: Why 3-Bit Breaks Compact Vision Models}
\label{sec:intro}

Quantization to 4-bit and 3-bit weights is widely used in large language models, where dense linear projections accumulate dot products across thousands of input channels and average out per-weight rounding noise~\citep{xiao2023smoothquant}. Compact computer vision backbones such as MobileNet and EfficientNet~\citep{tan2019efficientnet} are structured differently. To minimize multiply-accumulate operations on edge devices, they factor standard convolutions into $1\times 1$ pointwise channel projections and \textbf{$3\times 3$ or $5\times 5$ depthwise spatial convolutions}~\citep{jacob2018quantization,nagel2019dfq,li2021brecq}.

A $3\times 3$ depthwise filter contains only \textbf{9 scalar weights per channel}, with no cross-channel accumulation~\citep{nagel2019dfq,wang2019haq,li2021brecq}. When those 9 floating-point weights are quantized to a uniform 3-bit signed integer grid ($\{-3, -2, -1, 0, +1, +2, +3\}$, 7 levels) or a 4-bit grid ($\{-7, \dots, +7\}$, 15 levels), three degradation mechanisms compound:
\begin{enumerate}
  \item \textbf{Spatial Filter Degradation in Low-Channel Early Stages:} In the input stem and Stages 1--2 of EfficientNet-B0, the network has only 16 to 32 channels to extract spatial edges, corners, and color gradients from the $224\times 224$ input image. As demonstrated in mixed-precision architecture search~\citep{wang2019haq,shen2021oqat} and block-reconstruction post-training quantization (BRECQ)~\citep{li2021brecq}, early depthwise separable blocks exhibit much stronger second-order output sensitivity and lower structural redundancy than late pointwise projections: rounding a 9-weight edge detector to 7 discrete levels alters its spatial orientation before deeper layers process the feature map.
  \item \textbf{Outlier-Stretched Step Sizes:} Standard symmetric min-max quantization sets each channel's step size from its single largest magnitude weight: $\Delta = \max_i |w_i| / Q_{\max}$. As analyzed in minimum mean-squared error (OMSE) quantization~\citep{choukroun2019omse}, analytical 4-bit clipping~\citep{banner2019posttraining4bit}, and outlier-aware quantization~\citep{xiao2023smoothquant}, bell-shaped weight and activation distributions face a sharp trade-off between outlier clipping distortion and inner-bin rounding resolution. For 3-bit quantization ($Q_{\max}=3$), a single outlier weight $3\times$ larger than the channel's standard deviation stretches $\Delta$ wide enough that every weight with $|w_i| < 0.5\,\Delta$ rounds to the zero bin ($q_i = 0$).
  \item \textbf{Downstream BatchNorm Moment Drift:} Coarse weight rounding systematically shifts the mean and variance of each convolution's output activations~\citep{nagel2019dfq,banner2019posttraining4bit,li2021brecq}. When a quantized filter feeds a BatchNorm layer~\citep{jacob2018quantization} whose running moments were frozen during FP32 training, the shifted activations are mis-centered and either saturate the ReLU6 upper bound or collapse to zero.
\end{enumerate}

In this study, we quantify each stage of this failure cascade on ImageNet-1k~\citep{deng2009imagenet} and evaluate a four-stage recovery pipeline: \textbf{(i) early-layer W8 protection}, \textbf{(ii) per-channel MSE outlier clipping}~\citep{choukroun2019omse,banner2019posttraining4bit}, \textbf{(iii) 1,000-image BatchNorm recalibration}~\citep{nagel2019dfq}, and \textbf{(iv) 82-layer dynamic-programming (DP) bit allocation}~\citep{wang2019haq} \textbf{followed by quantization-aware training (QAT)}~\citep{zhuang2018lowbit,esser2020lsq,lee2025transitions}. Alongside our accuracy analysis, we benchmark exported FlatBuffer and packed-bitstream artifacts on CPU to evaluate the systems-level deployment trade-off between sub-byte storage savings and runtime unpacking latency.

\begin{figure}[!htbp]
\centering
\resizebox{\linewidth}{!}{%
\begin{tikzpicture}[
  font=\sffamily,
  >=Latex,
  mechbox/.style={
    draw=#1!80!black,
    fill=#1!8,
    rounded corners=4pt,
    line width=0.9pt,
    inner sep=0pt,
    minimum width=3.68cm,
    minimum height=2.60cm
  }
]

\node[anchor=west, font=\sffamily\bfseries\small, text=black!85] at (0, 6.45)
  {A. Why Uniform 3-Bit Quantization Fails on Compact CNNs (72.86\% $\rightarrow$ 0.30\%)};

\node[mechbox=red, anchor=north west] (m1) at (0, 6.05) {%
  \parbox[t][2.22cm][t]{3.34cm}{\footnotesize\raggedright
    \textbf{\small\textcolor{red!75!black}{1. $3\times 3$ Depthwise Filters}}\\[3pt]
    Only \textbf{9 weights/channel} with no cross-channel averaging; 7 levels ($\{-3..+3\}$) erase edge orientation.
  }%
};

\node[mechbox=orange!90!red, anchor=north west] (m2) at (4.06, 6.05) {%
  \parbox[t][2.22cm][t]{3.34cm}{\footnotesize\raggedright
    \textbf{\small\textcolor{orange!75!black}{2. Outlier Step Sizes}}\\[3pt]
    One outlier stretches max-abs step $\Delta = \max|w|/3$, collapsing inner weights into the zero bin ($q=0$).
  }%
};

\node[mechbox=orange, draw=orange!80!black, fill=yellow!12, anchor=north west] (m3) at (8.12, 6.05) {%
  \parbox[t][2.22cm][t]{3.34cm}{\footnotesize\raggedright
    \textbf{\small\textcolor{orange!80!black}{3. BatchNorm Drift}}\\[3pt]
    Weight rounding shifts channel means and variances, causing frozen FP32 BatchNorm moments to misfire.
  }%
};

\draw[->, line width=1.2pt, draw=red!70!black] (m1.east) -- (m2.west);
\draw[->, line width=1.2pt, draw=orange!80!black] (m2.east) -- (m3.west);

\node[anchor=west, font=\sffamily\bfseries\small, text=black!85] at (0, 2.95)
  {B. Five-Step Mixed-Precision Recovery Waterfall on EfficientNet-B0 (50k Validation)};

\draw[line width=0.6pt, draw=black!35] (3.35, -1.05) -- (8.98, -1.05);
\foreach \p/\x in {0/3.35, 20/4.85, 40/6.35, 60/7.85, 75/8.98} {
  \draw[line width=0.5pt, draw=black!35] (\x, -1.05) -- (\x, -1.17);
  \node[font=\sffamily\footnotesize, text=black!75, anchor=north] at (\x, -1.18) {\p\%};
}

\draw[dashed, line width=1.0pt, draw=blue!75!black] (8.81, -1.05) -- (8.81, 2.32);
\node[font=\sffamily\footnotesize\bfseries, text=blue!75!black, anchor=south] at (8.81, 2.32)
  {FP32: 72.86\% (INT8: 72.81\%)};

\node[anchor=east, font=\sffamily\footnotesize\bfseries, text=black!85] at (3.22, 2.05) {0. Uniform W3 (3.00b)};
\fill[red!80!black, rounded corners=1.5pt] (3.35, 1.87) rectangle (3.42, 2.23);
\node[anchor=west, font=\sffamily\footnotesize\bfseries, text=red!80!black] at (3.52, 2.05)
  {0.30\% (random chance)};

\node[anchor=east, font=\sffamily\footnotesize\bfseries, text=black!85] at (3.22, 1.47) {1. + Early W8 (0.34\%)};
\fill[orange!85!red, rounded corners=1.5pt] (3.35, 1.29) rectangle (3.86, 1.65);
\node[anchor=west, font=\sffamily\footnotesize\bfseries, text=orange!80!black] at (3.96, 1.47)
  {6.78\% (65.95\% if rest W4)};

\node[anchor=east, font=\sffamily\footnotesize\bfseries, text=black!85] at (3.22, 0.89) {2. + Tiered \& MSE Clip};
\fill[orange!80!yellow, rounded corners=1.5pt] (3.35, 0.71) rectangle (7.86, 1.07);
\node[anchor=east, font=\sffamily\footnotesize\bfseries, text=white] at (7.76, 0.89) {60.15\%};
\node[anchor=west, font=\sffamily\footnotesize\bfseries, text=black!80] at (9.02, 0.89)
  {3.55b (+9.00 pp clip)};

\node[anchor=east, font=\sffamily\footnotesize\bfseries, text=black!85] at (3.22, 0.31) {3. + 1K BN Recal};
\fill[teal!70!green!85!black, rounded corners=1.5pt] (3.35, 0.13) rectangle (8.16, 0.49);
\node[anchor=east, font=\sffamily\footnotesize\bfseries, text=white] at (8.06, 0.31) {64.09\%};
\node[anchor=west, font=\sffamily\footnotesize\bfseries, text=teal!60!black] at (9.02, 0.31)
  {3.55b (+3.94 pp BN)};

\node[anchor=east, font=\sffamily\footnotesize\bfseries, text=black!85] at (3.22, -0.27) {4. + 82-Layer DP};
\fill[cyan!65!blue!85!black, rounded corners=1.5pt] (3.35, -0.45) rectangle (8.45, -0.09);
\node[anchor=east, font=\sffamily\footnotesize\bfseries, text=white] at (8.35, -0.27) {67.94\%};
\node[anchor=west, font=\sffamily\footnotesize\bfseries, text=cyan!50!blue!85!black] at (9.02, -0.27)
  {3.56b (+3.85 pp DP)};

\node[anchor=east, font=\sffamily\footnotesize\bfseries, text=black!85] at (3.22, -0.85) {5. + 1-Ep QAT + BN};
\fill[blue!75!black, rounded corners=1.5pt] (3.35, -1.03) rectangle (8.52, -0.67);
\node[anchor=east, font=\sffamily\footnotesize\bfseries, text=white] at (8.42, -0.85) {68.98\%};
\node[anchor=west, font=\sffamily\footnotesize\bfseries, text=blue!80!black] at (9.02, -0.85)
  {3.56b ($-3.88$ pp gap)};

\end{tikzpicture}%
}
\caption{\textbf{Mechanisms of sub-4-bit quantization failure and the 5-step mixed-precision recovery waterfall on EfficientNet-B0.} \textbf{(A)} Uniform 3-bit quantization fails due to three compounding effects: (1) each $3\times 3$ depthwise spatial filter has only 9 weights per channel (unlike dense layers that average hundreds of inputs), so 7 signed levels ($\{-3..+3\}$) erase edge orientation in low-channel early stages; (2) a single outlier weight stretches the symmetric max-abs step size $\Delta = \max|w|/3$, collapsing the inner 95\% of weights into the zero bin ($q=0$); and (3) coarse weight rounding shifts output channel means and variances, causing frozen FP32 BatchNorm statistics to mis-center activations and saturate ReLU6. \textbf{(B)} Protecting $0.34\%$ of early weights ($14\text{ KB}$) at W8 rescues $+42.92$ pp when the rest of the model is W4 ($65.95\%$). For a $3.55$--$3.56$-bit mixed W8/W4/W3 model, combining per-channel MSE clipping ($+9.00$ pp), 1,000-image BatchNorm recalibration ($+3.94$ pp), 82-layer dynamic-programming bit allocation ($+3.85$ pp to $67.94\%$ PTQ), and 1 epoch of QAT with post-QAT BN recalibration recovers top-1 accuracy from $0.30\%$ to $68.98\%$ across all 50,000 validation images.}
\label{fig:collapse-and-recovery-ladder}
\end{figure}
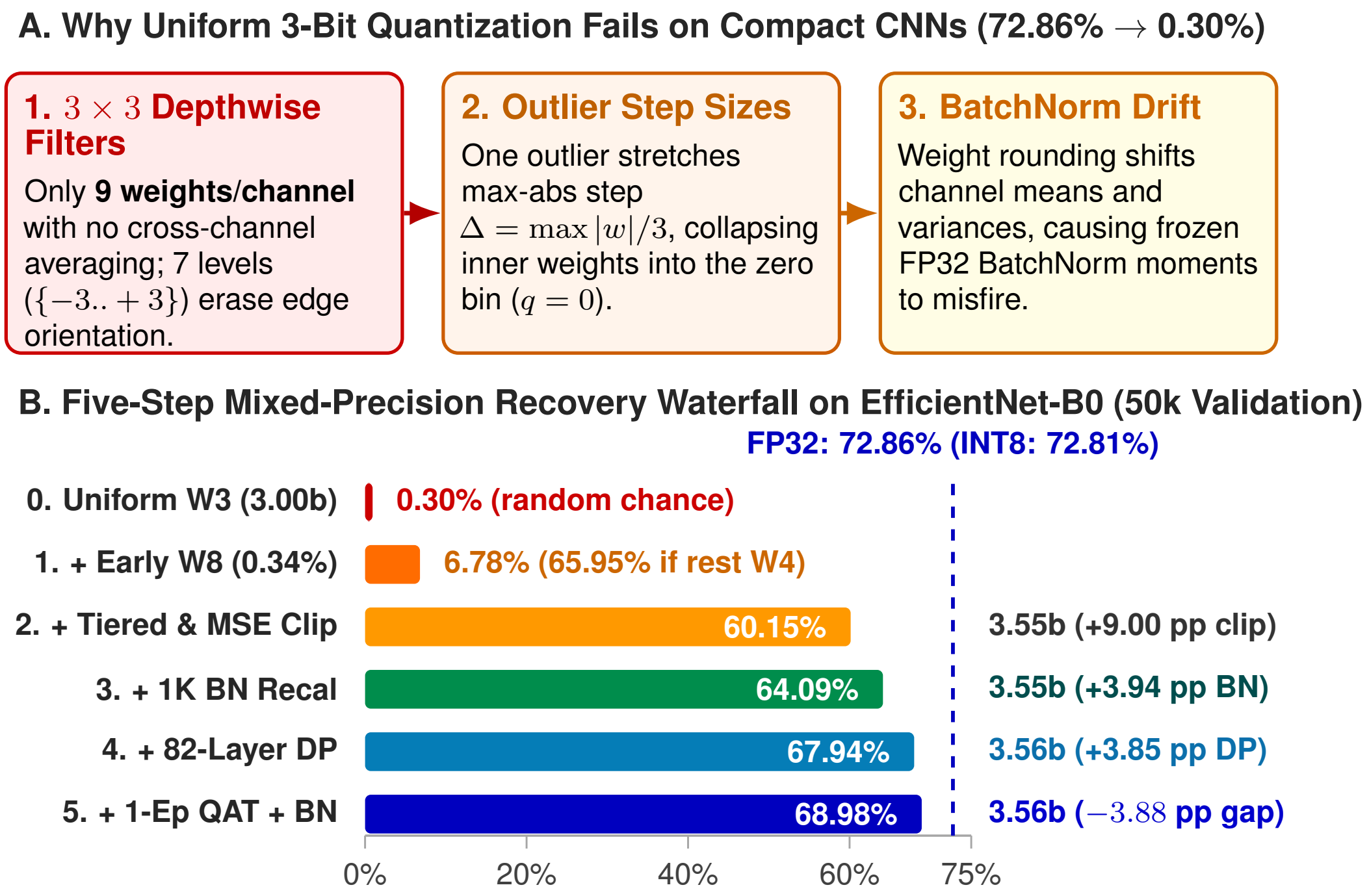

\section{Experimental Foundation: Baseline Checkpoints and Maturity Effects}
\label{sec:baselines-maturity}

To ensure reproducible, leakage-free comparisons, all evaluations in this paper use the complete 50,000-image ImageNet-1k validation set under verified deterministic preprocessing. Our reference backbones are scratch-trained ReLU6/SGD EfficientNet-B0, B1, and B2 models~\citep{tan2019efficientnet} trained for 88 epochs. Specifically, \textbf{EfficientNet-B0} ($5.32\text{M}$ parameters, $224\times 224$) reaches \textbf{72.864\%} top-1 ($91.106\%$ top-5) at its selected epoch-86 checkpoint ($72.840\%$ at terminal epoch 88); \textbf{EfficientNet-B1} ($7.85\text{M}$ parameters, $240\times 240$) reaches \textbf{74.908\%} top-1 ($92.192\%$ top-5) at epoch 87 ($74.826\%$ at epoch 88); and \textbf{EfficientNet-B2} ($9.17\text{M}$ parameters, $260\times 260$) reaches \textbf{75.952\%} top-1 ($92.752\%$ top-5) at epoch 88.

\subsection{Why Early-Epoch Pilots Underestimate Quantization Fragility}
\begin{figure}[!htbp]
\centering
\includegraphics[width=0.96\linewidth]{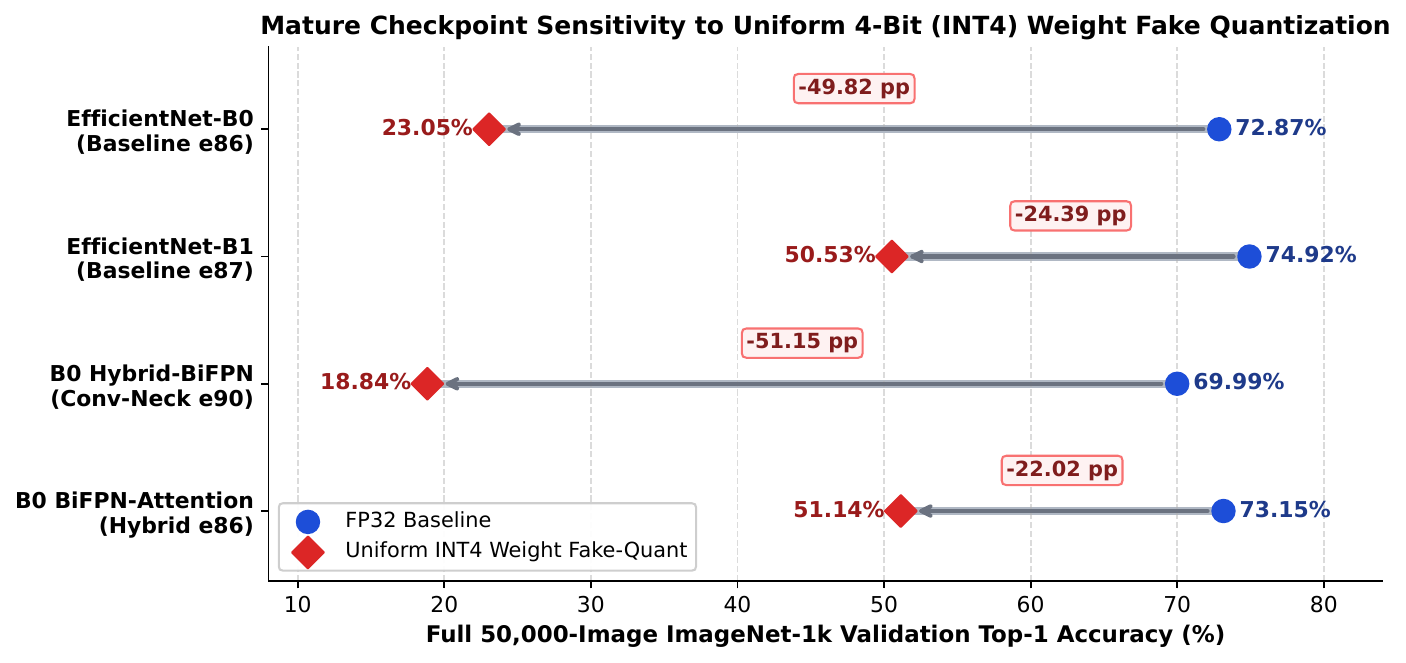}
\caption{\textbf{Uniform 4-bit (INT4) weight quantization on mature B0 and B1 checkpoints.} Converged checkpoints suffer severe post-training accuracy drops ($-49.82$ pp on B0 and $-24.39$ pp on B1) across all 50,000 validation images, contrasting sharply with early-epoch pilots.}
\label{fig:int4-sensitivity}
\end{figure}

A common pitfall in rapid quantization screening is evaluating low-bit schemes on lightly trained checkpoints. In our initial epoch-1 pilots ($5.88\%$--$18.88\%$ FP32 top-1), uniform INT4 and linear INT3 quantization appeared nearly lossless, losing less than $1.6$ percentage points. Once checkpoints converged above $70\%$ top-1, however, uniform post-training quantization degraded substantially (Figure~\ref{fig:int4-sensitivity}): under uniform W4 weight fake quantization, mature \textbf{B0 lost 49.82 points} ($72.86\% \rightarrow 23.03\%$), mature \textbf{B1 lost 24.39 points} ($74.92\% \rightarrow 50.53\%$), and a mature hybrid-BiFPN B0 checkpoint lost $51.15$ points ($69.99\% \rightarrow 18.84\%$). Early checkpoints rely on coarse, low-frequency representations that tolerate heavy rounding, whereas converged networks encode fine-grained decision boundaries across coupled depthwise and pointwise filters that break under uniform low-bit discretization~\citep{nagel2019dfq,li2021brecq,shen2021oqat}.

\subsection{Instability of Uniform 3-Bit QAT and Logarithmic Grids}
Before introducing mixed precision, we tested whether quantization-aware training (QAT)~\citep{jacob2018quantization,zhuang2018lowbit,zhuang2022progressive,esser2020lsq} could recover accuracy in a uniform 3-bit model. As analyzed by \citet{lee2025transitions}, latent weight updates in QAT only modify the discrete quantized model when crossing quantizer transition thresholds, making discrete weight transition rates sensitive to when fake quantization is enabled relative to the learning rate schedule. Comparing immediate linear-INT3 QAT against delayed linear-INT3 QAT (activating fake quantization halfway through an 80,072-step continuation from the mature B0 checkpoint across two random seeds) revealed pronounced seed-dependent variance. On \textbf{Seed 20261012}, the FP32 control scored $71.486\%$, immediate INT3 QAT reached $64.060\%$, and delayed INT3 QAT reached $64.424\%$ ($+0.364$ percentage points in favor of delay). On \textbf{Seed 20261013}, the FP32 control scored $71.444\%$, immediate INT3 QAT reached $65.242\%$, and delayed INT3 QAT fell to $54.612\%$ ($-10.630$ points penalty for delay).

We also evaluated multiplicative logarithmic quantization~\citep{lee2017lognet,oh2021logscale}. While base-2 logarithmic representations enable shift-add arithmetic in custom accelerators~\citep{lee2017lognet}, \citet{oh2021logscale} showed that single-word 3-bit logarithmic grids leave wide quantization gaps at the outer tail of the weight distribution unless paired with two-word sparse-tailed encoding (STLQ). Consistent with their analysis, our single-word factor-four logarithmic INT3 grid underperformed uniform linear INT3 by $2.92$ to $3.72$ points in controlled 5-arm pilots ($8.44\%/7.23\%$ vs.\ $11.37\%/10.95\%$). Together, these results indicate that uniform 3-bit quantization leaves EfficientNet-B0 $6.2$ to $16.8$ points below FP32 even after a full epoch of QAT.

\begin{figure}[!htbp]
\centering
\resizebox{\linewidth}{!}{%
\begin{tikzpicture}[
  font=\sffamily,
  >=Latex,
  tierbox/.style={
    draw=#1!80!black,
    fill=#1!8,
    rounded corners=4pt,
    line width=0.95pt,
    inner sep=0pt,
    minimum width=3.68cm,
    minimum height=4.15cm
  },
  stepbox/.style={
    draw=#1!80!black,
    fill=#1!8,
    rounded corners=4pt,
    line width=0.85pt,
    inner sep=0pt,
    minimum width=3.68cm,
    minimum height=2.45cm
  }
]

\node[anchor=west, font=\sffamily\bfseries\small, text=black!85] at (0, 6.45)
  {82-Layer Mixed-Precision Blueprint on EfficientNet-B0 (3.56 Mean Bits, 2.35 MB)};

\node[tierbox=blue, anchor=north west] (t0) at (0, 6.05) {%
  \parbox[t][3.75cm][t]{3.34cm}{\footnotesize\raggedright
    \textbf{\small\textcolor{blue!80!black}{Tier 0: Early Anchor}}\\[2pt]
    \textbf{Precision: W8} (0.34\%, 14 KB)\\[4pt]
    $\bullet$ Stem \& Stage 1--2 MBConv\\[2pt]
    $\bullet$ Stage 1 W3 drop: \textbf{$-47.62$ pp}\\[2pt]
    $\bullet$ Restores \textbf{+42.92 pp} on W4\\[2pt]
    $\bullet$ W8 $\rightarrow$ FP16: $+0.03$ pp
  }%
};

\node[tierbox=teal, anchor=north west] (t1) at (4.06, 6.05) {%
  \parbox[t][3.75cm][t]{3.34cm}{\footnotesize\raggedright
    \textbf{\small\textcolor{teal!80!black}{Tier 1: Spatial \& Head}}\\[2pt]
    \textbf{Precision: W4} ($\sim$55.8\% wts)\\[4pt]
    $\bullet$ Mid/deep $3\times 3$ \& $5\times 5$ DW\\[2pt]
    $\bullet$ Squeeze-and-Excitation (SE)\\[2pt]
    $\bullet$ Classifier ($1280\times 1000$)\\[2pt]
    $\bullet$ Sensitive $1\times 1$ projections
  }%
};

\node[tierbox=orange!90!red, anchor=north west] (t2) at (8.12, 6.05) {%
  \parbox[t][3.75cm][t]{3.34cm}{\footnotesize\raggedright
    \textbf{\small\textcolor{orange!80!black}{Tier 2: High-Fan-In Bulk}}\\[2pt]
    \textbf{Precision: W3} ($\sim$43.9\% wts)\\[4pt]
    $\bullet$ Wide $1\times 1$ pointwise convs\\[2pt]
    $\bullet$ Stages 5--7 expansion/proj\\[2pt]
    $\bullet$ High channel fan-in\\[2pt]
    $\bullet$ Assigned by 82-layer DP
  }%
};

\draw[->, line width=1.2pt, draw=blue!70!black] (t0.east) -- (t1.west);
\draw[->, line width=1.2pt, draw=teal!70!black] (t1.east) -- (t2.west);

\node[stepbox=purple, anchor=north west] (c1) at (0, 1.45) {%
  \parbox[t][2.05cm][t]{3.34cm}{\footnotesize\raggedright
    \textbf{\small\textcolor{purple!80!black}{Step A: MSE Clipping}}\\[3pt]
    Grid search $\alpha \in [0.50, 1.00]$ per channel to minimize $L_2$ rounding error (\textbf{+9.00 pp}).
  }%
};

\node[stepbox=teal, anchor=north west] (c2) at (4.06, 1.45) {%
  \parbox[t][2.05cm][t]{3.34cm}{\footnotesize\raggedright
    \textbf{\small\textcolor{teal!80!black}{Step B: 1K BN Recal}}\\[3pt]
    Updates BatchNorm moments on 1,000 training images to correct drift (\textbf{64.09\%} tiered).
  }%
};

\node[stepbox=blue, anchor=north west] (c3) at (8.12, 1.45) {%
  \parbox[t][2.05cm][t]{3.34cm}{\footnotesize\raggedright
    \textbf{\small\textcolor{blue!80!black}{Step C: DP + 1-Ep QAT}}\\[3pt]
    Knapsack DP ($\text{DP}_{0.5}$) reaches \textbf{67.94\% PTQ}; 1-ep QAT + BN reaches \textbf{68.98\%}.
  }%
};

\draw[->, line width=1.2pt, draw=purple!70!black] (c1.east) -- (c2.west);
\draw[->, line width=1.2pt, draw=teal!70!black] (c2.east) -- (c3.west);

\end{tikzpicture}%
}
\caption{\textbf{The 82-layer mixed-precision allocation blueprint for EfficientNet-B0 ($\text{DP}_{0.5}$, $3.56$ mean bits, $2.35\text{ MB}$).} \textbf{Top row:} Architectural precision tiers showing Tier~0 ($0.34\%$ of weights, $13,952$ parameters in the stem and Stage 1--2 depthwise/pointwise layers protected at W8, which alone restores $+42.92$ pp on W4 while FP16 adds only $+0.03$ pp), Tier~1 ($\sim 55.8\%$ of weights in mid/deep $3\times 3$ and $5\times 5$ depthwise filters, Squeeze-and-Excitation bottlenecks, and the $1280\times 1000$ classifier head assigned to W4), and Tier~2 ($\sim 43.9\%$ of weights in wide Stage 5--7 $1\times 1$ pointwise projections where high channel fan-in tolerates W3). \textbf{Bottom row:} The three post-selection calibration stages: per-output-channel MSE clipping ($\alpha^* = \arg\min_{\alpha \in [0.5, 1.0]} \|w - Q_\alpha(w)\|_2^2$), 1,000-image BatchNorm moment recalibration, and 82-layer dynamic-programming allocation followed by 1 epoch of QAT and post-QAT BN recalibration ($68.98\%$ top-1).}
\label{fig:layerwise-precision-blueprint}
\end{figure}
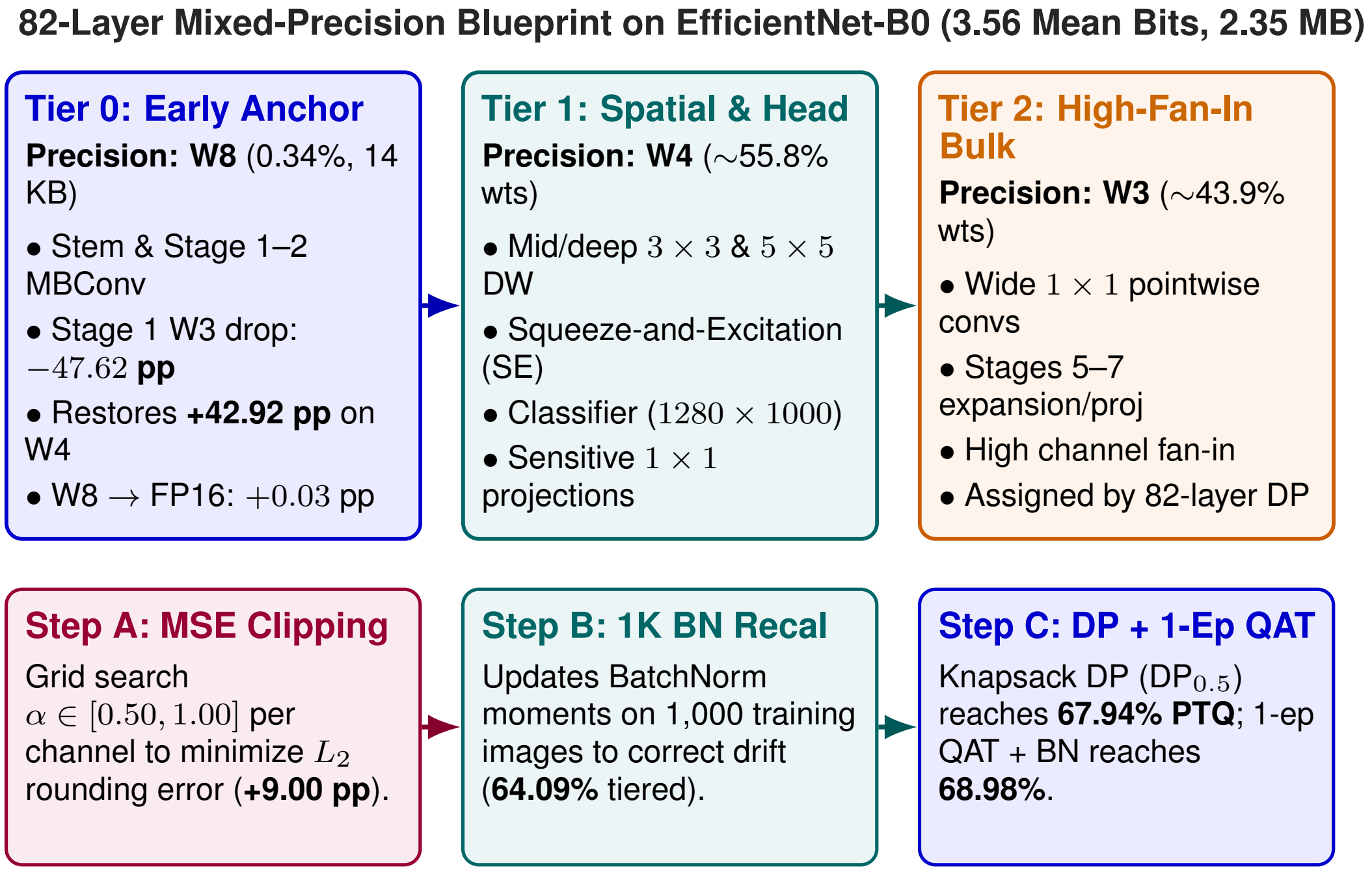

\section{Anatomy of Degradation: 82-Layer Sensitivity and Early-W8 Protection}
\label{sec:layerwise-sensitivity}

To identify which layers drive uniform W4 and W3 degradation, we conducted a 46-arm localized perturbation screen on a class-balanced 2,000-image validation subset (quantizing single stages or operator families while keeping the rest of EfficientNet-B0 in FP32) and verified the six most sensitive masks on the disjoint 48,000-image complement and the full 50,000-image validation set (Table~\ref{tab:layerwise-sensitivity}).

\begin{table}[!htbp]
\centering
\caption{\textbf{Localized vs.\ whole-model quantization sensitivity on mature EfficientNet-B0 (50,000-image validation).} Protecting $0.34\%$ of weights ($13,952$ parameters in early stem and Stage 1--2 depthwise/pointwise layers) at W8 restores $+42.92$ percentage points when the remaining $99.66\%$ of weights are quantized to W4.}
\label{tab:layerwise-sensitivity}
\footnotesize
\setlength{\tabcolsep}{4.0pt}
\resizebox{\linewidth}{!}{%
\begin{tabular}{@{}llrrr@{}}
\toprule
\textbf{Experiment Group} & \textbf{Quantized Scope / Precision Profile} & \textbf{Mean Bits} & \textbf{Top-1 (\%)} & \textbf{$\Delta$ vs.\ FP32 (pp)} \\
\midrule
\textbf{Reference} & Unquantized FP32 Baseline (B0 e86) & 32.00 & \textbf{72.87} & $0.00$ \\
\midrule
\textbf{Single-Mask} & Stage 1 Only at W3 (rest FP32) & 31.99 & 25.25 & \textbf{$-47.62$} \\
\textbf{Perturbations} & Early Depthwise Only (Stem + Stages 1--2) at W3 & 31.97 & 31.20 & \textbf{$-41.67$} \\
\textbf{(50k Verified)} & Early Depthwise Only (Stem + Stages 1--2) at W4 & 31.97 & 40.66 & $-32.21$ \\
 & Stage 1 Only at W4 (rest FP32) & 31.99 & 42.96 & $-29.91$ \\
 & Stage 2 Only at W3 (rest FP32) & 31.90 & 48.88 & $-23.99$ \\
 & Stage 2 Only at W4 (rest FP32) & 31.91 & 66.81 & $-6.06$ \\
\midrule
\textbf{Whole-Model} & Uniform W4 (All 82 layers at 4-bit) & 4.00 & 23.03 & $-49.83$ \\
\textbf{Profiles} & Uniform W3 (All 82 layers at 3-bit) & 3.00 & 0.30 & $-72.56$ \\
\textbf{(Max-Abs PTQ)} & \textbf{Early W8 ($0.34\%$ weights) + Rest W4 ($99.66\%$)} & \textbf{4.01} & \textbf{65.95} & \textbf{$-6.91$ ($+42.92$ vs.\ W4)} \\
 & Early W8 ($0.34\%$ weights) + Rest W3 ($99.66\%$) & 3.02 & 6.78 & $-66.09$ \\
 & Broad Heuristic W8/W4/W3 (Stages 4,7,Head at W3) & 3.51 & 51.92 & $-20.94$ \\
 & Narrow Heuristic W8/W4/W3 (Head + Classifier at W3) & 3.69 & 64.36 & $-8.51$ \\
\bottomrule
\end{tabular}%
}
\end{table}

As shown in Table~\ref{tab:layerwise-sensitivity} and Figure~\ref{fig:layerwise-precision-blueprint}, the primary bottleneck in uniform low-bit quantization lies in the earliest spatial layers. Quantizing \emph{only} Stage~1 ($0.03\%$ of weights) to W3 while keeping $99.97\%$ of the model in FP32 reduces top-1 accuracy by \textbf{47.62 percentage points}, and quantizing only the early depthwise filters (Stem and Stages 1--2) to W3 reduces accuracy by \textbf{41.67 points}. Conversely, protecting the stem and Stage 1--2 depthwise and pointwise layers at W8—just $13,952$ weights ($0.34\%$ of the model, or $14\text{ KB}$)—lifts whole-model W4 accuracy from \textbf{23.03\% to 65.95\%} ($+42.92$ points). To test whether these early layers require floating-point precision beyond 8 bits, we promoted Tier~0 from W8 to 16-bit floating point (FP16) across all 50,000 validation images; top-1 accuracy changed by only \textbf{$+0.026$ percentage points} ($64.118\%$ vs.\ $64.092\%$, paired 95\% CI $[-0.037, +0.089]$) while adding $58.7\text{ KB}$, confirming that 8-bit integer precision is sufficient for early spatial filtering.

At the same time, Table~\ref{tab:layerwise-sensitivity} shows why early-layer protection alone does not resolve 3-bit degradation across the rest of the network. When the remaining $99.66\%$ of weights are quantized to W3 under standard max-abs scaling, Early W8 protection raises accuracy only from $0.30\%$ to $6.78\%$. Recovering 3-bit capacity across middle and late inverted-residual blocks requires addressing inter-layer block sensitivity~\citep{li2021brecq}, outlier-stretched quantization scales~\citep{choukroun2019omse,banner2019posttraining4bit,xiao2023smoothquant}, BatchNorm moment drift~\citep{nagel2019dfq}, and layer-specific bit allocation~\citep{wang2019haq}.

\section{Rescuing Sub-4-Bit Models: Clipping, BatchNorm, DP Allocation, and QAT}
\label{sec:factorial-dp-qat}

To separate the effects of minimum-MSE weight clipping~\citep{choukroun2019omse,banner2019posttraining4bit}, BatchNorm statistics~\citep{nagel2019dfq}, and per-layer bit-width assignment~\citep{wang2019haq,li2021brecq}, we first measured quantization sensitivity across all 82 weighted layers at $\{W_3, W_4, W_8, \text{FP16}\}$ ($82 \times 4 = 328$ evaluations) under per-output-channel MSE grid clipping ($\alpha^* = \arg\min_{\alpha \in [0.5, 1.0]} \|w - Q_\alpha(w)\|_2^2$). Using an additive surrogate model of layerwise accuracy drops, we solved a dynamic program (DP) to find optimal 82-layer W8/W4/W3 assignments under three compression budgets ($\text{DP}_{0.5}$ at $3.56$ mean bits, $\text{DP}_{1.0}$ at $3.49$ bits, and $\text{DP}_{2.0}$ at $3.40$ bits).

We then executed a \textbf{16-arm full-validation factorial study} crossing four precision profiles (Heuristic Tiered Map, $\text{DP}_{0.5}$, $\text{DP}_{1.0}$, $\text{DP}_{2.0}$) $\times$ two scaling rules (Max-Abs vs.\ Per-Channel MSE Clipping) $\times$ two BatchNorm regimes (Stored FP32 Source BN vs.\ Recalibrated BN on a class-balanced 1,000-image training subset), followed by a 1-epoch QAT continuation on the $\text{DP}_{0.5}$ profile (Table~\ref{tab:factorial-qat}).

\begin{table}[!htbp]
\centering
\caption{\textbf{Full 50,000-image ImageNet-1k results for the 16-arm Precision $\times$ Clipping $\times$ BatchNorm factorial study and 1-epoch DP QAT continuation.} MSE clipping and 1,000-image BN recalibration exhibit a strong $-5.07$ pp interaction that lifts the 3.55-bit tiered map by $+12.94$ pp overall ($51.15\% \rightarrow 64.09\%$). Dynamic programming ($\text{DP}_{0.5}$, $3.56$ bits) adds $+3.85$ pp in PTQ ($67.94\%$) and reaches \textbf{68.98\%} after 1-epoch QAT + BN recalibration.}
\label{tab:factorial-qat}
\footnotesize
\setlength{\tabcolsep}{3.5pt}
\resizebox{\linewidth}{!}{%
\begin{tabular}{@{}lrrrrrrr@{}}
\toprule
\textbf{Precision Profile} & \textbf{Bits} & \textbf{Size} & \textbf{MaxAbs+Src} & \textbf{MSE+Src} & \textbf{MaxAbs+BN} & \textbf{MSE+BN} & \textbf{+1Ep QAT+BN} \\
\midrule
FP32 Reference (B0 e86) & 32.00 & 20.43 MB & 72.86\% & --- & 70.52\% & --- & 71.53\% (ctrl) \\
Uniform W8 Reference & 8.00 & 5.29 MB & 72.81\% & --- & 70.50\% & --- & --- \\
\midrule
Heuristic Tiered W8/W4/W3 & 3.55 & 2.35 MB & 51.15\% & 60.15\% & 60.15\% & 64.09\% ($+12.9$) & --- \\
\textbf{$\text{DP}_{0.5}$ Adaptive W8/W4/W3} & \textbf{3.56} & \textbf{2.35 MB} & 64.61\% & 66.22\% & 66.40\% & \textbf{67.94\%} ($+3.3$) & \textbf{68.98\%} ($+4.4$) \\
$\text{DP}_{1.0}$ Adaptive W8/W4/W3 & 3.49 & 2.30 MB & 63.30\% & 64.86\% & 65.81\% & 67.34\% ($+4.0$) & --- \\
$\text{DP}_{2.0}$ Adaptive W8/W4/W3 & 3.40 & 2.24 MB & 61.51\% & 63.37\% & 64.57\% & 66.35\% ($+4.8$) & --- \\
\bottomrule
\end{tabular}%
}
\end{table}

The factorial results in Table~\ref{tab:factorial-qat} demonstrate how outlier clipping, BatchNorm recalibration, and layerwise bit allocation interact during recovery. On the heuristic tiered map, per-channel MSE clipping alone~\citep{choukroun2019omse,banner2019posttraining4bit} improves accuracy by $+9.004$ points ($51.150\% \rightarrow 60.154\%$) and 1,000-image BN recalibration alone~\citep{nagel2019dfq} improves accuracy by $+9.004$ points ($51.150\% \rightarrow 60.154\%$), whereas combining both reaches $64.092\%$ ($+12.942$ points total; paired interaction $-5.066$ points, 95\% CI $[-5.531, -4.601]$). This sub-additive interaction occurs because clipping extreme weights reduces the largest activation variance shifts that otherwise cause frozen BN moments to mis-scale features, while BN recalibration corrects the remaining mean and variance drift.

Comparing bit-allocation strategies at matched model size ($3.56$ vs.\ $3.55$ bits, both $2.35\text{ MB}$), 82-layer dynamic programming ($\text{DP}_{0.5}$) outperforms the stage-level heuristic map by \textbf{$+3.848$ percentage points} under MSE clipping and recalibrated BN ($67.940\%$ vs.\ $64.092\%$). Consistent with block-sensitive mixed-precision analyses~\citep{wang2019haq,li2021brecq}, DP retains sensitive mid-network depthwise filters at W8 or W4 while assigning wide late $1\times 1$ projections to W3. Finally, fine-tuning $\text{DP}_{0.5}$ with one epoch of QAT (80,072 gradient updates) yields $66.072\%$ when evaluated with training-time EMA BatchNorm statistics because discrete quantizer step crossings~\citep{lee2025transitions} introduce batch-to-batch moment noise; applying a 1,000-image post-QAT BN recalibration pass lifts accuracy by \textbf{$+2.912$ points} (95\% CI $[+2.606, +3.218]$) to \textbf{68.984\%}, outperforming same-profile PTQ ($67.940\%$) by $+1.044$ points (95\% CI $[+0.796, +1.292]$). Even after all four interventions, however, a \textbf{3.88-point accuracy gap} remains relative to the $72.864\%$ FP32 baseline ($2.54$ points relative to the $71.528\%$ 1-epoch FP32 QAT control).

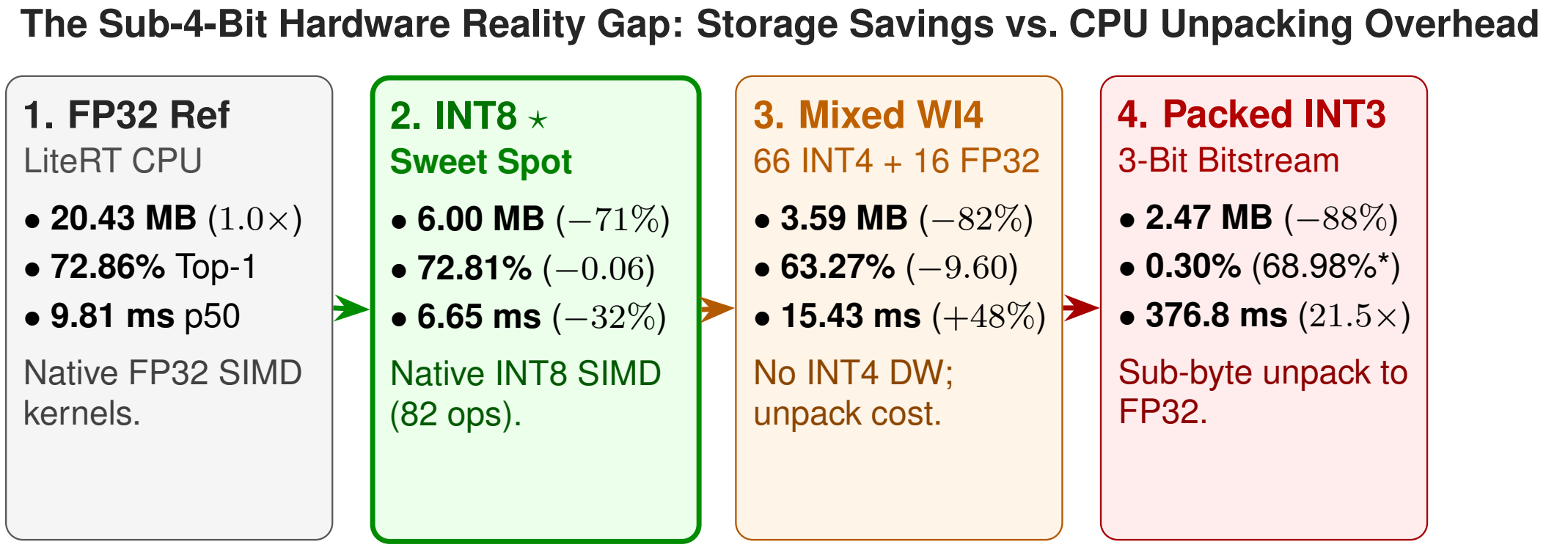
\begin{figure}[!htbp]
\centering
\resizebox{\linewidth}{!}{%
\begin{tikzpicture}[
  >=Stealth,
  font=\sffamily\footnotesize,
  hwcard/.style={draw=black!65, rounded corners=4pt, inner sep=0pt, minimum width=2.71cm, minimum height=3.85cm}
]

\node[anchor=west, font=\sffamily\small\bfseries, text=black!85] at (0, 4.35)
  {The Sub-4-Bit Hardware Reality Gap: Storage Savings vs.\ CPU Unpacking Overhead};

\node[hwcard, fill=gray!8, draw=gray!70!black, anchor=north west] (c1) at (0, 3.95) {%
  \parbox[t][3.45cm][t]{2.42cm}{\footnotesize\raggedright
    \textbf{\small\color{black!85}1. FP32 Ref}\\[1pt]
    \textcolor{black!70}{LiteRT CPU}\\[4pt]
    $\bullet$ \textbf{20.43 MB} ($1.0\times$)\\[2pt]
    $\bullet$ \textbf{72.86\%} Top-1\\[2pt]
    $\bullet$ \textbf{9.81 ms} p50\\[4pt]
    \textcolor{black!75}{Native FP32 SIMD kernels.}
  }%
};

\node[hwcard, fill=green!8, draw=green!55!black, very thick, anchor=north west] (c2) at (3.03, 3.95) {%
  \parbox[t][3.45cm][t]{2.42cm}{\footnotesize\raggedright
    \textbf{\small\color{green!45!black}2. INT8 $\star$}\\[1pt]
    \textbf{\color{green!50!black}Sweet Spot}\\[4pt]
    $\bullet$ \textbf{6.00 MB} ($-71\%$)\\[2pt]
    $\bullet$ \textbf{72.81\%} ($-0.06$)\\[2pt]
    $\bullet$ \textbf{6.65 ms} (\textbf{$-32\%$})\\[4pt]
    \textcolor{green!35!black}{Native INT8 SIMD (82 ops).}
  }%
};

\node[hwcard, fill=orange!9, draw=orange!70!black, anchor=north west] (c3) at (6.06, 3.95) {%
  \parbox[t][3.45cm][t]{2.42cm}{\footnotesize\raggedright
    \textbf{\small\color{orange!75!black}3. Mixed WI4}\\[1pt]
    \textcolor{orange!70!black}{66 INT4 + 16 FP32}\\[4pt]
    $\bullet$ \textbf{3.59 MB} ($-82\%$)\\[2pt]
    $\bullet$ \textbf{63.27\%} ($-9.60$)\\[2pt]
    $\bullet$ \textbf{15.43 ms} (\textbf{$+48\%$})\\[4pt]
    \textcolor{orange!55!black}{No INT4 DW; unpack cost.}
  }%
};

\node[hwcard, fill=red!7, draw=red!65!black, anchor=north west] (c4) at (9.09, 3.95) {%
  \parbox[t][3.45cm][t]{2.42cm}{\footnotesize\raggedright
    \textbf{\small\color{red!70!black}4. Packed INT3}\\[1pt]
    \textcolor{red!65!black}{3-Bit Bitstream}\\[4pt]
    $\bullet$ \textbf{2.47 MB} ($-88\%$)\\[2pt]
    $\bullet$ \textbf{0.30\%} (68.98\%*)\\[2pt]
    $\bullet$ \textbf{376.8 ms} (\textbf{$21.5\times$})\\[4pt]
    \textcolor{red!55!black}{Sub-byte unpack to FP32.}
  }%
};

\draw[->, very thick, draw=green!55!black] (c1.east) -- (c2.west);
\draw[->, very thick, draw=orange!70!black] (c2.east) -- (c3.west);
\draw[->, very thick, draw=red!65!black] (c3.east) -- (c4.west);

\end{tikzpicture}%
}
\caption{\textbf{The deployment gap between storage compression and execution speed on EfficientNet-B0.} Converting from 32-bit floats ($20.43\text{ MB}$, $9.81\text{ ms}$) to full-integer INT8 ($6.00\text{ MB}$, $6.65\text{ ms}$) simultaneously reduces file size by $70.7\%$ and median CPU latency by $32.2\%$ with negligible accuracy loss ($-0.06$ pp) because all 82 convolutional and dense layers execute on native 8-bit integer SIMD instructions. Below 8 bits, standard CPU runtimes lack native sub-byte depthwise kernels: LiteRT's mixed 4-bit path (WI4, 66 INT4 + 16 FP32 depthwise operators) and custom packed 3-bit (INT3) bitstreams reduce storage to $3.59\text{ MB}$ ($-82.4\%$) and $2.47\text{ MB}$ ($-88.4\%$), but runtime bit-unpacking into floating-point buffers slows inference to $15.43\text{ ms}$ ($+47.9\%$) and $376.8\text{ ms}$ ($21.5\times$ slower than FP32; $0.30\%$ uniform INT3 accuracy in the packed CPU benchmark vs.\ *$68.98\%$ for the $3.56$-bit $\text{DP}_{0.5}+\text{QAT}$ recovery model).}
\label{fig:hardware-reality-gap}
\end{figure}

\section{The Deployment Reality Gap: Storage vs.\ CPU Latency}
\label{sec:deployment-pareto}

\begin{figure}[!htbp]
\centering
\includegraphics[width=0.96\linewidth]{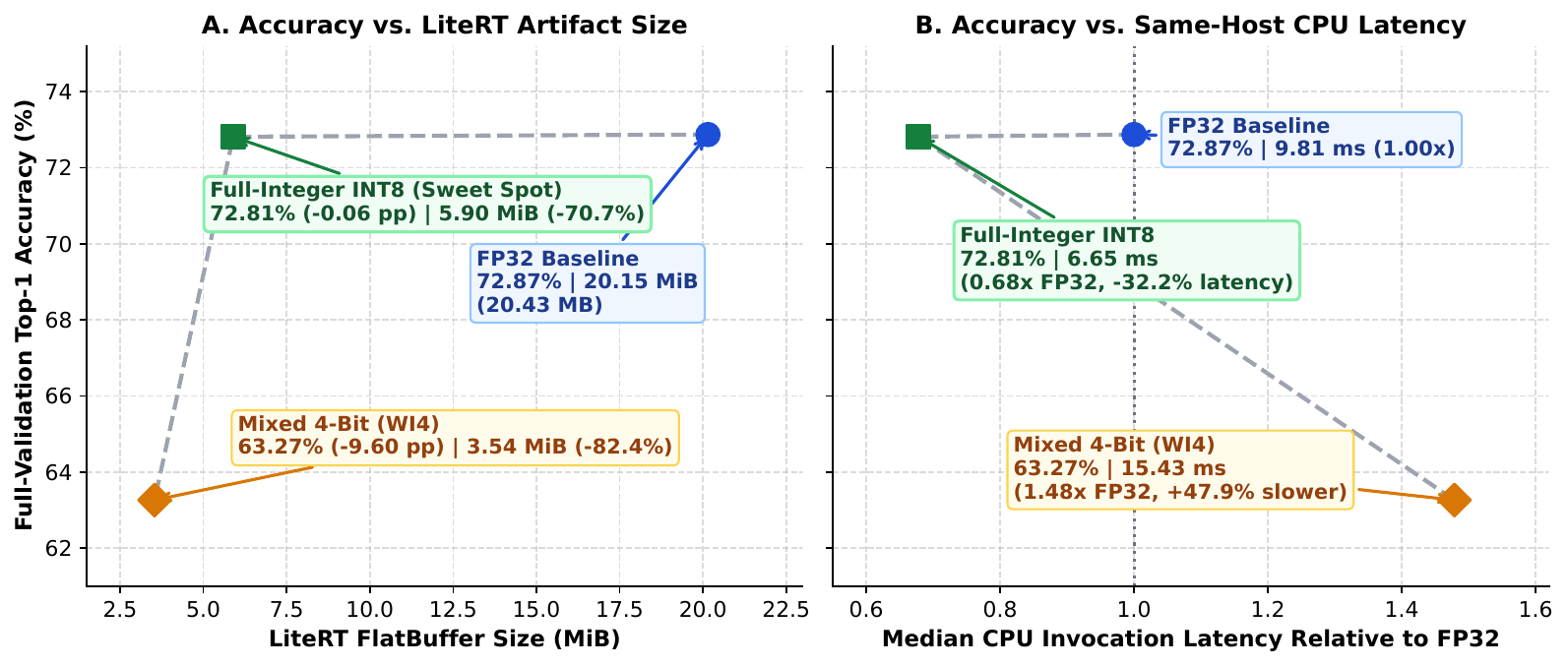}
\caption{\textbf{Measured LiteRT artifact size, full-validation accuracy, and CPU latency on mature EfficientNet-B0.} Full-integer INT8 dominates FP32 on both size and speed with negligible accuracy loss ($-0.058$ pp). Mixed 4-bit (WI4) reduces storage further ($-82.4\%$) but loses $9.60$ points and increases median latency by $47.9\%$ due to missing native sub-byte depthwise kernels.}
\label{fig:deployment-pareto}
\end{figure}

Simulated fake quantization measures representational quality inside an FP32 compute graph, but practical edge deployment depends on whether the target runtime provides native low-bit arithmetic kernels~\citep{jacob2018quantization,wang2019haq,nvidia_tensorrt_114}. To evaluate this systems boundary directly, we benchmarked exported FlatBuffer and packed-weight artifacts on CPU across all 50,000 validation images (Table~\ref{tab:deployment-benchmarks} and Figure~\ref{fig:deployment-pareto}).

\begin{table}[!htbp]
\centering
\caption{\textbf{Real-world CPU artifact size, accuracy, and invocation latency on mature EfficientNet-B0 (50,000-image validation).} Full-integer INT8 defines the practical hardware Pareto frontier ($-70.7\%$ size, $-32.2\%$ latency, $-0.06$ pp accuracy). Sub-4-bit formats reduce disk and RAM footprint further but execute slower on CPU due to bit-unpacking overhead. Note that the Packed Linear INT3 row benchmarks a uniform 3-bit bitstream ($0.30\%$ top-1, $2.47\text{ MB}$) to isolate raw 3-bit CPU unpacking latency ($376.8\text{ ms}$), whereas the $3.56$-bit $\text{DP}_{0.5}+\text{QAT}$ recovery model reaches $68.98\%$ top-1 ($2.35\text{ MB}$, Table~\ref{tab:factorial-qat} and Figure~\ref{fig:hardware-reality-gap}) under simulated quantization.}
\label{tab:deployment-benchmarks}
\footnotesize
\setlength{\tabcolsep}{3.5pt}
\resizebox{\linewidth}{!}{%
\begin{tabular}{@{}llrrrrl@{}}
\toprule
\textbf{Runtime Path} & \textbf{Precision Format} & \textbf{Size} & \textbf{Size $\Delta$} & \textbf{Top-1} & \textbf{CPU p50 / p95 (ms)} & \textbf{Execution Mechanism} \\
\midrule
LiteRT (TFLite) & Float32 Baseline & 20.43 MB & $0.0\%$ & 72.86\% & 9.81 / 10.98 & Native FP32 SIMD kernels \\
\textbf{LiteRT (TFLite)} & \textbf{Full-Integer INT8} & \textbf{6.00 MB} & \textbf{$-70.7\%$} & \textbf{72.81\%} & \textbf{6.65 / 7.22 ($-32.2\%$)} & \textbf{Native INT8 SIMD (82 ops)} \\
LiteRT (TFLite) & Mixed 4-Bit (WI4) & 3.59 MB & $-82.4\%$ & 63.27\% & 15.43 / 16.43 ($+47.9\%$) & 66 INT4 + 16 FP32 DW; unpack \\
Custom C++/TF & Packed Linear INT3 (Unif.) & 2.47 MB & $-88.4\%$ & 0.30\% & 376.8 / 386.6 ($21.5\times$) & Bitstream unpack to FP32 \\
\bottomrule
\end{tabular}%
}
\end{table}

As shown in Figures~\ref{fig:hardware-reality-gap} and~\ref{fig:deployment-pareto}, full-integer INT8 LiteRT~\citep{jacob2018quantization} maps all 82 weighted layers—both $1\times 1$ pointwise projections and $3\times 3$/$5\times 5$ depthwise convolutions—directly onto hardware-accelerated 8-bit integer vector instructions (such as \texttt{SDOT}/\texttt{UDOT} in ARMv8-A NEON and \texttt{VPDPBUSD} in x86 AVX2/AVX-512 VNNI, which execute four 8-bit multiply-accumulates per lane per cycle). Consequently, INT8 simultaneously shrinks the FlatBuffer artifact by $70.7\%$ ($20.43\text{ MB} \rightarrow 6.00\text{ MB}$) and lowers median CPU latency by $32.2\%$ on B0 ($9.81\text{ ms} \rightarrow 6.65\text{ ms}$) and by $13.5\%$ on Hybrid-BiFPN B0 ($11.88\text{ ms} \rightarrow 10.28\text{ ms}$) with minimal accuracy loss ($-0.058$ pp and $-0.568$ pp).

Below 8 bits, however, theoretical bit-width savings diverge sharply from execution speed due to microarchitectural constraints on standard ARM and x86 CPUs. Commodity CPU vector ISAs expose byte-aligned register files and 8-bit dot-product datapaths, but lack native sub-byte (4-bit or 3-bit) vector multiply-accumulate instructions (e.g., there is no 4-bit equivalent of \texttt{SDOT} in ARMv8-A or x86 VNNI) and provide no hardware support for non-byte-aligned 3-bit addressing. Furthermore, mainstream inference runtimes restrict optimized INT4 weight-only kernels to high-arithmetic-intensity dense GEMM/matmul layers rather than low-reuse $3\times 3$/$5\times 5$ depthwise convolutions~\citep{wang2019haq,nvidia_tensorrt_114}. Executing sub-byte convolutions on general-purpose CPUs therefore requires software bit-manipulation—bitwise shifts, nibble masking, sign-extension, and buffer repacking—to expand packed 4-bit or 3-bit weights into 8-bit or 32-bit registers prior to arithmetic execution. In LiteRT's mixed 4-bit weight path (WI4, which quantizes 66 pointwise/dense layers to INT4 while leaving 16 depthwise layers in FP32), file size drops to $3.59\text{ MB}$ ($-82.4\%$), but on-the-fly software unpacking increases median CPU latency to $15.43\text{ ms}$ ($+47.9\%$ slower than FP32 and $2.32\times$ slower than INT8). In our custom packed-INT3 bitstream benchmark ($2.47\text{ MB}$, $-88.4\%$ weight storage), unpacking non-byte-aligned 3-bit signed integers back into floating-point buffers requires $376.8\text{ ms}$ per image ($21.5\times$ slower than FP32). Translating the $3.56$-bit storage footprint of our $68.98\%$ mixed-precision model into wall-clock speedup therefore requires pairing mixed-precision bit allocation with dedicated sub-byte lookup-table (LUT) SIMD kernels or specialized low-bit NPU hardware.

\section{Scope, Limitations, and Conclusion}
\label{sec:conclusion}

\paragraph{Methodological Scope.}
All accuracy comparisons in Sections~\ref{sec:layerwise-sensitivity} and~\ref{sec:factorial-dp-qat} use deterministic 50,000-image ImageNet-1k evaluation with fake-quantized weights in TensorFlow FP32 operators, isolating representational quantization error from hardware kernel implementation. Real packed storage and CPU latency are measured separately in Section~\ref{sec:deployment-pareto}. Our B0/B1/B2 reference checkpoints follow an 88-epoch ReLU6/SGD baseline protocol ($3.4$--$3.9$ points below the 350--400-epoch Swish/RMSProp baselines of \citet{tan2019efficientnet}); cross-scale comparisons are descriptive because batch sizes differed across recovery runs, and the $\text{DP}_{0.5}$ QAT continuation represents a single-seed, single-epoch budget.

\paragraph{Conclusion.}
Sub-4-bit quantization in compact vision backbones is constrained by both representational sensitivity and systems-level execution support. On the representational side, uniform 3-bit and 4-bit quantization suffer severe degradation because 9-weight depthwise spatial stencils in early layers lack cross-channel redundancy against coarse rounding and outlier-stretched step sizes~\citep{li2021brecq,choukroun2019omse,banner2019posttraining4bit}. By protecting $0.34\%$ of early weights ($14\text{ KB}$) at 8-bit, combining per-channel MSE clipping with 1,000-image BatchNorm recalibration~\citep{nagel2019dfq}, allocating layerwise bit-widths via dynamic programming ($3.56$ mean bits)~\citep{wang2019haq}, and running post-QAT BN recalibration, we recover top-1 accuracy on EfficientNet-B0 from $0.30\%$ to $68.98\%$. On the systems side, our CPU benchmarks highlight a critical \textbf{hardware reality gap} (Figures~\ref{fig:hardware-reality-gap} and~\ref{fig:deployment-pareto}): without native sub-byte convolutional kernels, sub-4-bit WI4 and packed INT3 models incur a $1.48\times$ to $21.5\times$ runtime unpacking penalty relative to FP32. For production deployment on standard CPUs, full-integer INT8 ($72.81\%$ top-1, $-70.7\%$ size, $-32.2\%$ latency) remains the Pareto-dominant configuration until sub-byte LUT or NPU kernels are paired with mixed-precision W8/W4/W3 allocation.

\makeatletter
\if@anonymous
\paragraph{Code and Weights Availability.}
Trained checkpoints, 82-layer dynamic-programming bit-allocation profiles, exported LiteRT FlatBuffers, and SHA-256 verification manifests will be released publicly upon publication.
\else
\paragraph{Code and Weights Availability.}
Trained checkpoints, 82-layer dynamic-programming bit-allocation profiles, exported LiteRT FlatBuffers, and SHA-256 verification manifests for this study (arXiv:2611.23445) are publicly available at \url{https://huggingface.co/meatloafer/efficientnet-b0-sub4bit-mixed-precision} (companion sparse-attention and weight-sharing study: arXiv:2611.23446, \url{https://huggingface.co/meatloafer/efficientnet-b0-stage-attention}).
\fi
\makeatother

\bibliographystyle{plainnat}
\bibliography{references}

\end{document}